\documentclass[letterpaper,10pt,conference]{ieeeconf}
\IEEEoverridecommandlockouts
\usepackage{amsmath,amssymb,amsfonts}
\usepackage[pdftex]{graphicx}
\usepackage{textcomp}
\usepackage{xcolor}
\usepackage{soul}
\usepackage{titletoc}

\usepackage[bookmarks=true,colorlinks=false,hidelinks]{hyperref}
\newcommand{\link}[2]{\textcolor{magenta}{\href{#1}{#2}}}

\let\labelindent\relax
\usepackage{enumitem}

\usepackage{float}
\usepackage{listings}
\usepackage{multirow}
\usepackage{xparse}
\usepackage{optidef}
\usepackage{algorithm}
\usepackage{algpseudocode}

\makeatletter
\let\NAT@parse\undefined
\makeatother
\usepackage[numbers,sort,compress]{natbib}

\usepackage[all=normal,bibbreaks=tight,paragraphs=tight,floats=tight]{savetrees}

\ifx\arxiv\undefined

\else

\fi

\newcommand{\obj}[1]{#1}

\NewDocumentCommand \proposition {g g g g} {\texttt{#1}(#2
  \IfValueTF{#3}{,\,#3}{}
  \IfValueTF{#4}{,\,#4}{}
  )
}

\NewDocumentCommand \action {>{\SplitList{ }} m} {\actioncall #1}
\NewDocumentCommand \actioncall {g g g g} {\text{#1}(
  \IfValueTF{#2}{#2}{}
  \IfValueTF{#3}{,#3}{}
  \IfValueTF{#4}{,#4}{}
  )
}

\DeclareMathOperator*{\argmax}{arg\,max}

\newcommand{\func}[2]{\mathop{}#1\left(#2\right)}
\newcommand{\prob}[1]{\func{p}{#1}}
\newcommand{\given}{\;\middle|\;}
\newcommand{\E}[2]{\mathop{}\operatorname{E}_{#1}\left[#2\right]}
\newcommand{\V}[2]{\mathop{}\operatorname{Var}_{#1}\left[#2\right]}

\lstdefinelanguage{pddl}{
    morekeywords={forall},
    otherkeywords={:goal},
    sensitive=true, %
    morecomment=[l]{;},
    morestring=[b]" %
}

\usepackage[font=footnotesize,justification=justified,singlelinecheck=false]{caption}
\usepackage{etoolbox}
\makeatother
\def\tablename{Table}

\title{
    \LARGE \bf STAP: Sequencing Task-Agnostic Policies \\
    \vspace{5pt}
    \normalsize{Project page: \link{https://sites.google.com/stanford.edu/stap/home}{sites.google.com/stanford.edu/stap}}
    \vspace{-11pt}
}

\author{
    \href{https://www.chrisagia.com/}{Christopher Agia*}, \href{https://cs.stanford.edu/~takatoki/}{Toki Migimatsu*}, \href{https://jiajunwu.com/}{Jiajun Wu}, \href{https://web.stanford.edu/~bohg/}{Jeannette Bohg}
    \thanks{*Authors contributed equally to this work.}
    \thanks{Toyota Research Institute provided funds to support this work.}%
    \vspace{2pt} \\
    \small{Department of Computer Science, Stanford University, California, U.S.A.} \\
    \small{\texttt{Email: \{\href{mailto:cagia@stanford.edu}{cagia},\href{mailto:takatoki@stanford.edu}{takatoki},\href{mailto:jiajunw@stanford.edu}{jiajunw},\href{mailto:bohg@stanford.edu}{bohg}\}@stanford.edu}}
    \vspace{-8pt}
}

\begin{document}
\maketitle

\begin{abstract}
Advances in robotic skill acquisition have made it possible to build general-purpose libraries of learned skills for downstream manipulation tasks. 
However, naively executing these skills one after the other is unlikely to succeed without accounting for dependencies between actions prevalent in long-horizon plans.
We present Sequencing Task-Agnostic Policies (STAP), a scalable framework for training manipulation skills and coordinating their geometric dependencies at planning time to solve long-horizon tasks never seen by any skill during training.
Given that Q-functions encode a measure of skill feasibility, we formulate an optimization problem to maximize the joint success of all skills sequenced in a plan, which we estimate by the product of their Q-values.
Our experiments indicate that this objective function approximates ground truth plan feasibility and, when used as a planning objective, reduces myopic behavior and thereby promotes long-horizon task success.
We further demonstrate how STAP can be used for task and motion planning by estimating the geometric feasibility of skill sequences provided by a task planner. 
We evaluate our approach in simulation and on a real robot.
\end{abstract}

\section{Introduction}
\label{sec:intro}
\begin{figure}
    \centering
    \includegraphics[width=\columnwidth]{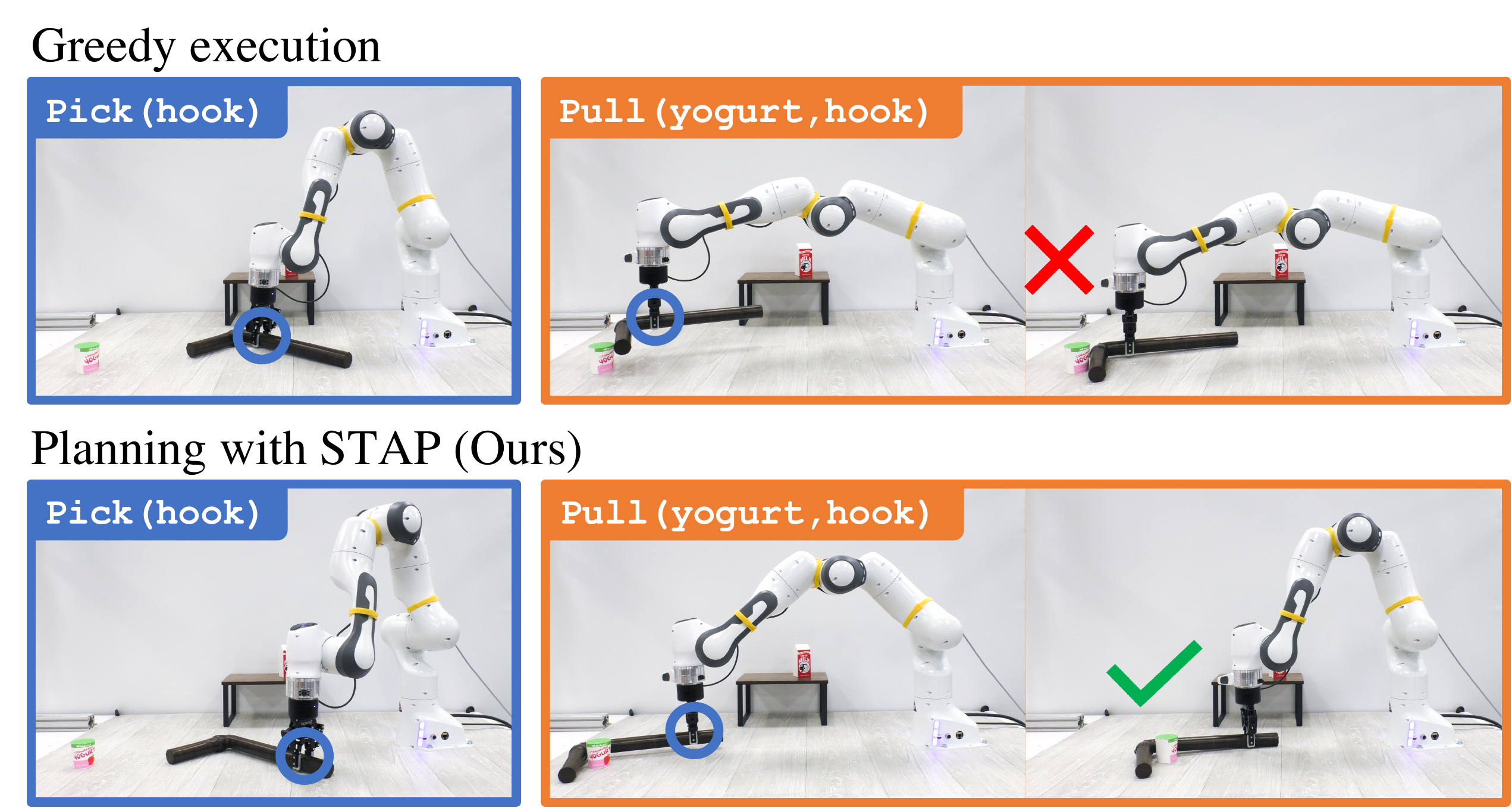}
    \caption{Sequential manipulation tasks often contain geometric dependencies between actions. In this example, the robot needs to use the hook to pull the block into its kinematic workspace so it is close enough to pick up. The top row shows how greedy execution of skills results in the robot picking up the hook in a way that prevents it from reaching the block. We present a method for planning with skills to maximize long-horizon success without the need to train the skills on long-horizon tasks.}
    \label{fig:teaser}
\end{figure}

Performing sequential manipulation tasks requires a robot to reason about dependencies between actions. Consider the example in Fig.~\ref{fig:teaser}, where the robot needs to retrieve an object outside of its workspace by first using an L-shaped hook to pull the target object closer. How the robot picks up the hook affects whether the target object will be reachable. %

Traditionally, planning actions to ensure the geometric feasibility of a sequential manipulation task is handled by motion planning~\cite{lavalle:2006,toussaint2015-lgp,toussaint2018differentiable}, which typically requires full observability of the environment state and knowledge of its dynamics. %
Learning-based approaches~\cite{Kaelbling1996ReinforcementLA,argall2009survey,Schaal1999-SCHIIL-2} can acquire skills without this privileged information. 
However, using independently learned skills to perform unseen long-horizon manipulation tasks is an unsolved problem. 
The skills could be myopically executed one after another to solve a simpler subset of tasks, but solving more complex tasks requires planning with these skills to ensure the feasibility of the entire skill sequence.

Prior work focuses on sequencing skills at \textit{train time} to solve a small set of sequential manipulation tasks~\cite{xu2021-daf, dalal2021-raps}. 
To contend with long-horizons, these methods often learn skills~\cite{da2012learning} that consist of a policy and parameterized manipulation primitive~\cite{felip2013manipulation}. The policy predicts the parameters of the primitive, thereby governing its motion.
Such methods are \textit{task-specific} in that they need to be trained on skill sequences that reflect the tasks they might encounter at test time.
In our framework, we assume that a task planner provides a novel sequence of skills at \textit{test time} that will then be grounded with parameters for manipulation primitives through optimization.
This makes our method \textit{task-agnostic}, as skills can be sequenced to solve long-horizon tasks not seen during training.

At the core of our method, {\em Sequencing Task-Agnostic Policies} (STAP), we use Q-functions to optimize the parameters of manipulation primitives in a given sequence.
Policies and Q-functions for each skill are acquired through off-the-shelf Reinforcement Learning. 
We then define a planning objective to maximize all Q-functions in a skill sequence, ensuring its geometric feasibility.
To evaluate downstream Q-functions of future skills, we learn a dynamics model that can predict future states. 
We also use {\em Uncertainty Quantification\/} (UQ) to avoid visiting states that are {\em Out-Of-Distribution\/} (OOD) for the learned skills. 
We train all of these components independently per skill, making it easy to gradually expand a library of skills without the need to retrain existing ones.

Our contributions are three-fold: we propose 
1) a framework to train an extensible library of task-agnostic skills, 
2) a planning method that optimizes arbitrary sequences of skills to solve long-horizon tasks,
and 3) a method to solve Task and Motion Planning (TAMP) problems with learned skills. 
In extensive experiments, we demonstrate that planning with STAP promotes long-horizon success on tasks with complex geometric dependencies between actions.
We also demonstrate that our framework works on a real robot.

\section{Related Work}
\label{sec:related}
\subsection{Robot skill learning}
How to represent and acquire composable manipulation skills is a widely studied problem in robotics. 
A broad class of methods uses Learning from Demonstration (LfD)~\cite{argall2009survey}. 
Dynamic Movement Primitives (DMPs) \cite{schaal2006-dmp,pastor2009-dmp,khansari2011-dmp} are a form of LfD that learns the parameters of dynamical systems encoding movements~\cite{ijspeert2002movement,matsubara2010learning,ude2010task,kulvicius2011joining}.
More recent extensions integrate DMPs with deep neural networks to learn more flexible policies~\cite{bahl2020-ndp,bahl2021-hndp}---for instance, to build a large library of skills from human video demonstrations~\cite{shao2021-concept2robot}.
Skill discovery methods instead identify action patterns in offline datasets~\cite{shankar2019-discovering} and either distill them into policies~\cite{shankar2020-learning,ajay2020-opal} or extract skill priors for use in downstream tasks~\cite{singh2020-parrot,pertsch2020-spiral}. 
Robot skills can also be acquired via active learning~\cite{da2014active}, Reinforcement Learning (RL)~\cite{rajeswaran2017-dext-manip,kalashnikov2018-qtopt,kalashnikov2021-mtopt,sharma2019-dynamics,lu2022-awopt}, and offline RL~\cite{chebotar2021-actionable}. 

An advantage of our planning framework is that it is agnostic to the types of skills employed, requiring only that it is possible to predict the probability of the skill's success given the current state and action.
Here, we learn skills~\cite{da2012learning} that consist of a policy and a parameterized manipulation primitive~\cite{felip2013manipulation}. 
The actions output by the policy are the parameters of the primitive determining its motion. 
In STAP, we will use the Q-functions of the policy to optimize suitable parameters~\cite{kalashnikov2018-qtopt,shao2021-concept2robot} for a sequence of manipulation primitives.

\subsection{Long-horizon robot planning}
Once manipulation skills have been acquired, using them to perform sequential manipulation tasks remains an open challenge.
\cite{kappler2015data,kaelbling2017learning,ames2018learning,migimatsu2022symbolic} propose data-driven methods to determine the \textit{symbolic} feasibility of skills and only control their timing, while we seek to ensure the \textit{geometric} feasibility of skills by controlling their trajectories.
Other techniques rely on task planning~\cite{wu2021-embr,huang2019-crsp}, subgoal planning~\cite{simeonov2020-subgoal-skills}, or meta-adaptation~\cite{xu2018-ntp, huang2019-ntg} to sequence learned skills to novel long-horizon goals.  
However, the tasks considered in these works do not feature rich geometric dependencies between actions that necessitate motion planning or skill coordination.

The options framework~\cite{sutton1999between} and the parameterized action
Markov Decision Process (MDP)~\cite{masson2016-pamdps} train a high-level policy to engage low-level policies~\cite{bacon2017-option,nachum2018-hiro} or primitives~\cite{dalal2021-raps,chitnis2020-efficient,vuong2021learning,nasiriany2022-augmenting} towards long-horizon goals.
\cite{shah2021-vfs} proposes a hierarchical RL method that uses the value functions of lower-level policies as the state space for a higher-level RL policy.
Our work is also related to model-based RL methods which jointly learn dynamics and reward models to guide planning~\cite{finn2017-dvf,chua2018-pets,hafner2019-planet}, policy search~\cite{janner2019-mbpo,hafner2019-dreamer}, or combine both~\cite{xie2020-skill,sekar2020-plan2explore}. 
While these methods demonstrate that policy hierarchies and model-based planning can enable RL to solve long-horizon problems, they are typically trained in the context of a single task.
In contrast, we seek to \textit{plan} with lower-level skills to solve tasks never seen before.

Closest in spirit to our work is that of \citet{xu2021-daf}, Deep Affordance Foresight (DAF), which proposes to learn a dynamics model, skill-centric affordances (value functions), and a skill proposal network that serves as a higher-level RL policy. 
We identify several drawbacks with DAF: 
first, because DAF relies on multi-task experience for training, generalizing beyond the distribution of training tasks may be difficult; 
second, the dynamics, affordance models, and skill proposal network need to be trained synchronously, which complicates expanding the current library of trained skills;
third, their planner samples actions from uniform random distributions, which prevents DAF from scaling to high-dimensional action spaces and long horizons. 
STAP differs in that our dynamics, policies, and affordances (Q-functions) are learned independently per skill. 
Without any additional training, we combine the skills at planning time to solve unseen long-horizon tasks. 
We compare our method against DAF in the planning experiments (Sec.~\ref{sec:experiments-planning}).

\subsection{Task and motion planning}

TAMP solves problems that require both symbolic and geometric reasoning~\cite{garrett2021integrated, toussaint2015-lgp}. 
DAF learns a skill proposal network to replace the typical task planner in TAMP, akin to~\cite{wang2022-gtp}.
Another prominent line of research learns components of the TAMP system, often from a dataset of precomputed solutions~\cite{wang2018-aml,driess2020-dvr,chitnis2020-camps,silver2021-ploi,driess2021-lgr,wang2021learning}.
The problems we consider involve complex geometric dependencies between actions that are typical in TAMP.
However, STAP only performs geometric reasoning and by itself is not a TAMP method. 
We demonstrate in experiments (Sec.~\ref{sec:experiments-tamp}) that STAP can be combined with symbolic planners to solve TAMP problems.

\section{Problem Setup}
\label{sec:problem}
\subsection{Long-horizon planning}
\label{sec:planning-tamp}

Our objective is to solve long-horizon manipulation tasks that require sequential execution of learned skills.
These skills come from a skill library $\mathcal{L} = \{\psi^1, \dots, \psi^K\}$, where each skill $\psi^k$ consists of a parameterized manipulation primitive~\cite{felip2013manipulation} $\phi^k$ and a learned policy $\pi^k$.
A primitive $\func{\phi^k}{a^k}$ takes in parameters $a^k$ and executes a series of motor commands on the robot, while a policy $\func{\pi^k}{a^k \given s^k}$ is trained to predict a distribution of suitable parameters $a^k$ from the current state $s^k$.
For example, the $\actioncall{Pick}{a}{b}$ skill may have a primitive which takes as input an end-effector pose and executes a trajectory to pick up object $\obj{a}$, where the robot first moves to the commanded pose, closes the gripper to grasp $\obj{a}$, and then lifts $\obj{a}$ off of $\obj{b}$.
The learned policy $\pi^k$ for this skill will then try to predict end-effector poses to pick up $\obj{a}$.

We assume access to a high-level planner that computes \textit{plan skeletons} (i.e. skill sequences) to achieve a high-level goal.
STAP aims to solve the problem of turning plan skeletons into geometrically feasible \textit{action plans} (i.e. parameters for each manipulation primitive in the plan skeleton).

STAP is agnostic to the choice of high-level planner. 
For instance, it can be used in conjunction with Planning Domain Definition Language (PDDL) \cite{mcdermott1998pddl} task planners to perform hierarchical TAMP~\cite{kaelbling2011hierarchical}.
In this setup, the task planner and STAP will be queried numerous times to find multiple plan skeletons grounded with optimized action plans.
STAP will also evaluate each action plan's probability of success (i.e. its geometric feasibility).
After some termination criterion is met, such as a timeout, the candidate plan skeleton and action plan with the highest probability of success is returned.

\subsection{Task-agnostic policies}

We aim to learn policies $\{\pi^1, \dots, \pi^K\}$ for the skill library $\mathcal{L}$ that can be sequenced by a high-level planner in arbitrary ways to solve any long-horizon task.
We call these policies task-agnostic because they are not trained to solve a specific long-horizon task.
Instead, each policy $\pi^k$ is associated with a skill-specific contextual bandit (i.e. a single timestep MDP)
\begin{align}\label{eq:skill-mdp}
    \mathcal{M}^k = \left(\mathcal{S}^k, \mathcal{A}^k, T^k, R^k, \rho^k\right),
\end{align}
where $\mathcal{S}^k$ is the state space, $\mathcal{A}^k$ is the action space, $\func{T^k\!}{s'^k \given s^k, a^k}$ is the transition model, $\func{R^k\!}{s^k, a^k, s'^k}$ is the binary reward function, and $\func{\rho^k\!}{s^k}$ is the initial state distribution.
Given a state $s^k$, the policy $\pi^k$ produces an action $a^k$, and the state evolves according to the transition model $\func{T^k\!}{s'^k \given s^k, a^k}$. 
Thus, the transition model encapsulates the execution of the manipulation primitive $\phi^k$ (Sec.~\ref{sec:planning-tamp}). 

A long-horizon domain is one in which each timestep involves the execution of a single policy, and it is specified by 
\begin{align}
    \overline{\mathcal{M}}
        &= \left(\mathcal{M}^{1:K}, \overline{\mathcal{S}}, \overline{T}^{1:K}, \overline{\rho}^{1:K}, \Gamma^{1:K}\right), \label{eq:long-horizon-domain}
\end{align}
where $\mathcal{M}^{1:K}$ is the set of MDPs whose policies can be executed in the long-horizon domain, $\overline{\mathcal{S}}$ is the state space of the long-horizon domain, $\func{\overline{T}^k\!}{\overline{s}' \given \overline{s}, a^k}$ is an extension of dynamics $\func{T^k\!}{s'^k \given s^k, a^k}$ that models how the entire long-horizon state evolves with action $a^k$, $\func{\overline{\rho}^k}{\overline{s}}$ is an extension of initial state distributions $\func{\rho^k}{s^k}$ over the long-horizon state space, and $\Gamma^k: \overline{\mathcal{S}} \rightarrow \mathcal{S}^k$ is a function that maps from the long-horizon state space to the state space of policy $k$. We assume that the dynamics $\func{T^k\!}{s'^k \given s^k, a^k}, \func{\overline{T}^k\!}{\overline{s}' \given \overline{s}, a^k}$ and initial state distributions $\func{\rho^k\!}{s^k}, \func{\overline{\rho}^k\!}{\overline{s}^k}$ are unknown.

Note that while the policies may have different state spaces $\mathcal{S}^k$, policy states $s^k$ must be obtainable from the long-horizon state space $\overline{\mathcal{S}}$ via $s^k = \func{\Gamma^k\!}{\overline{s}}$. 
This is to ensure that the policies can be used together in the same environment to perform long-horizon tasks. 
In the base case, all the state spaces are identical and $\Gamma^k$ is simply the identity function. 
Another case is that $\overline{s}$ is constructed as the concatenation of all $s^{1:K}$ and $\func{\Gamma^k\!}{\overline{s}}$ extracts the slice in $\overline{s}$ corresponding to $s^k$.

\section{Sequencing Task-Agnostic Policies}
\label{sec:taps}
Given a task in the form of a sequence of skills to execute, our planning framework constructs an optimization problem with the policies, Q-functions, and dynamics models of each skill.
Solving the optimization problem results in parameters for all manipulation primitives in the skill sequence such that the entire sequence's probability of success is maximized.

We formalize our planning methodology in this section and outline its implementation in Sec.~\ref{sec:planning}. 
Lastly, we describe our procedure for training modular skill libraries in Sec.~\ref{sec:training}.

\subsection{Grounding skill sequences with action plans}
\label{sec:taps-grounding}

\begin{figure}
    \centering
    \includegraphics[width=\columnwidth]{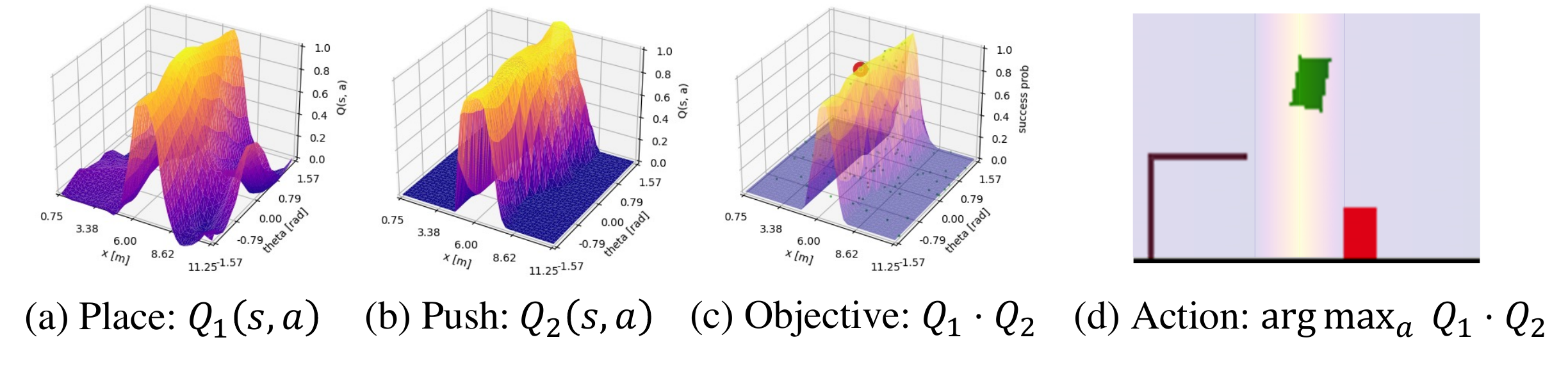}
    \caption{Planning in a 2D toy domain. The agent needs to get the green block under the brown receptacle with two skills: $\action{Place}$ and $\action{Push}$ that operate on the horizontal position $x$ of the green block.
    Plots (a) and (b) show the Q-functions across $(x, \theta)$ for each skill. 
    $\action{Place}$ is only trained to get the green block on the ground, so the planner must determine $a=x$ \textit{s.t.} $\action{Push}$ is unobstructed. 
    The optimal action maximizes the probability of long-horizon task success (Eq.~\ref{eq:planning-objective}), approximated by the product of Q-functions in plot (c).}
    \label{fig:pybox2d}
\end{figure}

We assume that we are given a plan skeleton of skills $\tau = [\psi_1, \dots, \psi_H] \in \mathcal{L}^H$ (hereafter denoted by $\tau = \psi_{1:H}$) that should be successfully executed to solve a long-horizon task.
Let $\mathcal{M}_h$ with subscript $h$ denote the MDP corresponding to the $h$-th skill in the sequence---in contrast to $\mathcal{M}^k$ with superscript $k$, which denotes the $k$-th MDP in the skill library. 
A long-horizon task is considered successful if every skill reward $r_1, \dots, r_H$ received during execution is $1$.

Given an initial state $\overline{s}_1 \in \overline{\mathcal{S}}$, our problem is to ground the plan skeleton $\tau = \psi_{1:H}$ with an action plan $\xi = [a_1, \dots, a_H] \in \mathcal{A}_1 \times \dots \times \mathcal{A}_H$ that maximizes the probability of succeeding at the long-horizon task. This is framed as an optimization problem $\argmax_{a_{1:H}} J$, where the maximization objective $J$ is the task success probability
\begin{equation*}
    J(a_{1:H}; \overline{s}_1)
        = \prob{r_1 = 1, \dots, r_H = 1 \given \overline{s}_1, a_{1:H}}.
\end{equation*}
Here, $r_{1:H}$ are the skill rewards received at each timestep.

With the long-horizon dynamics models $\func{\overline{T}^k\!}{\overline{s}' \given \overline{s}, a^k}$, the objective can be cast as the expectation
\begin{equation*}
    J%
        = \E{\overline{s}_{2:H} \sim \overline{T}_{1:H-1\!}\!}{\prob{r_1 = 1, \ldots, r_H = 1 \given \overline{s}_{1:H}, a_{1:H}}}.
\end{equation*}
By the Markov assumption, rewards are conditionally independent given states and actions. We can express the probability of task success as the product of reward probabilities
\begin{equation*}
    J%
        = \E{\overline{s}_{2:H} \sim \overline{T}_{1:H-1\!}\!}{\Pi_{h=1}^H \prob{r_h = 1 \given \overline{s}_h, a_h}}.
\end{equation*}
Because the skill rewards are binary, the skill success probabilities are equivalent to Q-values:
\begin{align*}
    \prob{r_h = 1 \given \overline{s}_h, a_h}
        &= \E{\overline{s}_{h+1} \sim \overline{T}_{h}}{r_h \given \overline{s}_h, a_h} \\
        &= \func{Q_h}{\!\func{\Gamma_h}{\overline{s}_h}, a_h}.
\end{align*}
The final objective is expressed in terms of Q-values:
\begin{equation}
    J%
        = \E{\overline{s}_{2:H} \sim \overline{T}_{1:H-1}}{\Pi_{h=1}^{H} \func{Q_h}{\!\func{\Gamma_h}{\overline{s}_h}, a_h}}. \label{eq:planning-objective}
\end{equation}

This planning objective is simply the product of Q-values evaluated along the trajectory $(\overline{s}_1, a_1, \dots, \overline{s}_H, a_H)$, where the states are predicted by the long-horizon dynamics model: $\overline{s}_2 \sim \func{\overline{T}_1}{\cdot \given \overline{s}_1, a_1}, \dots, \overline{s}_H \sim \func{\overline{T}_{H-1}}{\cdot \given \overline{s}_{H-1}, a_{H-1}}$.\footnote{One might consider maximizing the \textit{sum} of Q-values instead of the product, but this may not reflect the probability of task success. For example, if we want to optimize a sequence of ten skills, consider a plan that results in nine Q-values of $1$ and one Q-value of $0$, for a total sum of $9$. One Q-value of $0$ would indicate just one skill failure, but this is enough to cause a failure for the entire task. Compare this to a plan with ten Q-values of $0.9$. This plan has an equivalent sum of $9$, but it is preferable because it has a non-zero probability of succeeding.}

\subsection{Ensuring action plan feasibility}

A plan skeleton $\tau = \psi_{1:H}$ is feasible only if, for every pair of consecutive skills $\psi_i$ and $\psi_j$, there is a non-zero overlap between the terminal state distribution of $i$ and the initial state distribution of $j$. 
More formally,
\begin{equation}
    \E{\overline{s}_i \sim \overline{\rho}_i, a_i \sim \mathcal{A}_i, \overline{s}_j \sim \overline{\rho}_j}{\func{\overline{T}_i\!}{\overline{s}_j \given \overline{s}_i, a_i}}
        > 0, \label{eq:transition-feasibility}
\end{equation}
where $\overline{\rho}_i$ and $\overline{\rho}_j$ are the initial state distributions for skills $\psi_i$ and $\psi_j$, respectively, and $a_i$ is uniformly distributed with respect to action space $\mathcal{A}_i$ for skill $\psi_i$. 
Given a state $\overline{s}_i \sim \overline{\rho}_i$, it is part of the planner's job to determine an action $a_i$ that induces a valid subsequent state $\overline{s}_j \sim \overline{\rho}_j$ if one exists.
Failing to do so constitutes an OOD event for skill $\psi_j$, where the state $\overline{s}_j$ has drifted beyond the region of the state space where $\func{Q_j}{\func{\Gamma_j}{\overline{s}_j}, a_j}$ is well-defined and $\psi_j$ is executable.

Neglecting state distributional shift over an action plan $\xi$ may degrade the quality of objective function $J$ with spuriously high Q-values (Eq.~\ref{eq:planning-objective}).
Moreover, Eq.~\ref{eq:transition-feasibility} cannot be explicitly computed to determine the validity of actions because the initial state distributions of all skills $\func{\overline{\rho}^k\!}{\overline{s}^k}$ are unknown.
We can detect OOD states (and actions) by performing UQ on the Q-functions $Q^k(s^k, a^k)$. 
Filtering out Q-values with high uncertainty would result in action plans $\xi$ that are robust (i.e. have low uncertainty) while maximizing the task feasibility objective. We discuss efficient methods for training UQ models on learned Q-functions in Sec.~\ref{sec:training-scod}.

\begin{figure*}[htb!]
    \centering
    \begin{minipage}{0.42\textwidth}
        \centering
        \vspace{0.07cm}
        \includegraphics[width=\linewidth]{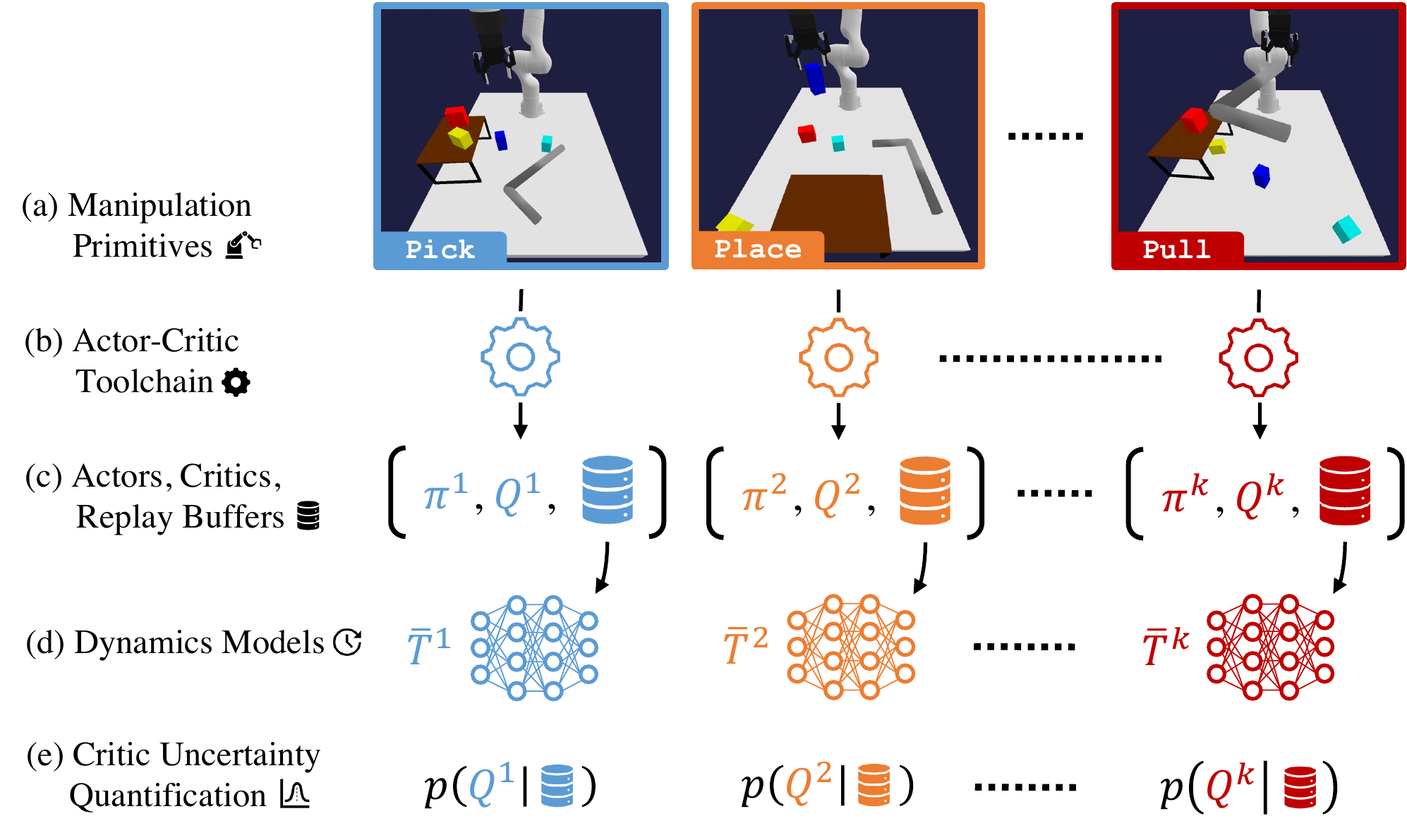}
        \label{fig:taps-training}
    \end{minipage}
    \hfill
    \begin{minipage}{0.56\textwidth}
        \centering
        \includegraphics[width=\linewidth]{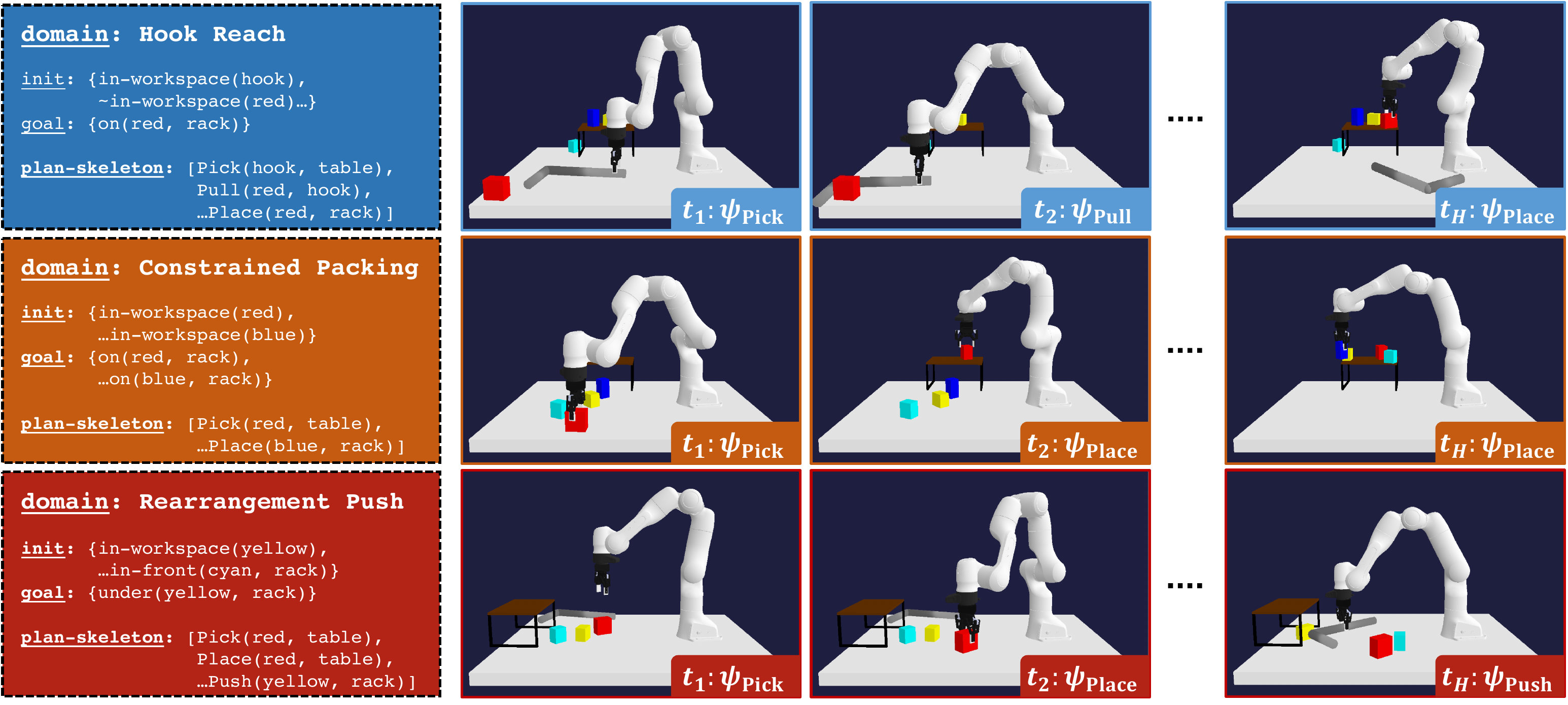}
        \label{fig:taps-tasks}
    \end{minipage}
    \vspace*{-3mm}
    \caption{\textbf{Left:} Training pipeline. We train each skill independently on single-step environments with skill-specific rewards. We use the experience collected by each skill to train (1) a dynamics model and (2) a SCOD~\cite{sharma2021-scod} model to predict OOD critic inputs per skill. The benefit of training each skill independently is that we can easily add skills to the library or even mix skill acquisition strategies (e.g. RL, imitation learning, and handcrafted skills). Our planning framework ensures that the skills can be composed to solve any long-horizon task even if the skills were not explicitly trained to perform those tasks. \textbf{Right:} Example evaluation tasks. We evaluate our method on 9 tasks: 3 \texttt{Hook Reach} tasks, where the robot needs to use the hook to bring objects closer, 3 \texttt{Constrained Packing} tasks, where the robot needs to place blocks on the rack, and 3 \texttt{Rearrangement Push} tasks, where the robot needs to remove obstacles to push a target block under the rack. The 9 tasks feature a range of geometric complexities and plan skeleton lengths.}
    \label{fig:taps-methods}
\end{figure*}

\section{Planning Action Sequences}
\label{sec:planning}

\label{sec:planning-actions}

To find action sequences that maximize the probability of long-horizon task success (Eq.~\ref{eq:planning-objective}), we use sampling-based optimization techniques: shooting and cross-entropy method (CEM)~\cite{rubinstein1999-cem}. In shooting, we simply sample action plans $\xi = a_{1:H} \in \mathcal{A}_1 \times \dots \times \mathcal{A}_H$ and select the one with the highest predicted objective score. 
CEM is an extension of shooting that iteratively refines the action sampling distribution to fit a fraction of the population with the highest objective scores.

Sampling action plans from uniform distributions may be sufficient for small action spaces and short skill sequences. 
However, this strategy suffers from the curse of dimensionality and may not scale desirably to the large action spaces and long skill sequences that we consider.
Meanwhile, directly executing actions $a^k \sim \func{\pi^k\!}{\cdot | s^k}$ from policies that are trained to solve skill-specific tasks produces myopic behavior that rarely succeeds for long-horizon tasks with complex geometric dependencies between actions.

The policies can be leveraged to initialize a sampling-based search by producing an action plan that is likely to be closer to an optimal plan than one sampled uniformly at random. 
We therefore use two variants of shooting and CEM, termed policy shooting and policy CEM, which sample actions from Gaussian distributions $a^k \sim \func{\mathcal{N}}{\func{\pi^k\!}{s^k}, \sigma}$, where the mean is the action predicted by the policy and the standard deviation is a planning hyperparameter.

\section{Training Skills}
\label{sec:training}
\subsection{Policies and Q-functions}
\label{sec:training-policies}

One of the key advantages of our approach is that the policies can be trained independently and then composed at test time to solve unseen sequential tasks. 
For each skill $\psi_k$, we want to obtain a policy $\pi^k: \mathcal{S}^k \rightarrow \mathcal{A}^k$ that solves the task specified by the skill-specific MDP $\mathcal{M}^k$ (Eq.~\ref{eq:skill-mdp}), along with a Q-function $Q^k$ modeling the policy's expected success
\begin{equation*}
    Q^k(s^k, a^k)
        = \E{s'^k \sim \func{T^k}{\cdot \given s^k, a^k}}{\!\func{R^k\!}{s^k, a^k, s'^k}}.
\end{equation*}

Our framework is agnostic to the method for acquiring the policy and Q-function. 
Many deep RL algorithms are able to simultaneously learn the policy (i.e. actor) and Q-function (i.e. critic) with unknown dynamics~\cite{lillicrap2015-ddpg,fujimoto2018-td3}. 
We therefore leverage off-the-shelf RL algorithms to learn a policy and Q-function for each skill (Fig.~\ref{fig:taps-methods} - Left (c)). 
In our experiments, we specifically use Soft Actor-Critic (SAC)~\cite{haarnoja2018-sac}.
For other policy acquisition methods, policy evaluation can be performed to obtain a Q-function after a policy has been learned.

\subsection{Dynamics}
\label{sec:training-dynamics}
The dynamics models are used to predict future states at which each downstream Q-function in the plan skeleton will be evaluated. 
We learn a deterministic model $\func{\overline{T}^k}{\overline{s}, a^k}$ for each skill $\psi_k$ using the single-step forward prediction loss
\begin{equation*}
    \func{L_\text{dynamics}}{\overline{T}^k; \overline{s}, a^k, \overline{s}'}
        = \left\| \func{\overline{T}^k\!}{\overline{s}, a^k} - \overline{s}' \right\|_2^2.
\end{equation*}
Each dynamics model $\overline{T}^k$ is trained on the state transition experience $(\overline{s}, a^k, \overline{s}')$ collected during the training of policy $\pi^k$ (Sec.~\ref{sec:training-policies}), stored in the replay buffer $\mathcal{D}^k$ (Fig.~\ref{fig:taps-methods} - Left (d)).
Training our dynamics models on existing state transitions is efficient and circumvents the challenges associated with learning dynamics in the context of a long-horizon task~\cite{janner2019-mbpo}.

\subsection{Uncertainty quantification}
\label{sec:training-scod}
Measuring the epistemic uncertainty over the Q-values allows us to identify when dynamics-predicted states and planned actions drift OOD for downstream critics $Q^k$.
We leverage recent advances in neural network UQ to obtain an explicit Gaussian posterior predictive distribution
\begin{equation}
    \prob{Q^k \given s^k, a^k, \mathcal{D}^k; w^k}
        = \func{\mathcal{N}}{\mu_{Q^k}, \sigma_{Q^k}; w^k}
    \label{eq:scod-posterior}
\end{equation}
with sketching curvature for OOD detection (SCOD)~\cite{sharma2021-scod}. 
SCOD computes the weights $w^k$ that parameterizes the posterior distribution over each critic $Q^k$ using only the experience $(\overline{s}, a^k) \sim \mathcal{D}^k$ collected over the course of training policy $\pi^k$ (Fig.~\ref{fig:taps-methods} - Left (e)).
An advantage of SCOD over common UQ techniques~\cite{ensemblereview2021,dropoutunc2016} is that it imposes no train-time dependencies on any algorithms used in our framework.

\section{Experiments}
\label{sec:experiments}
In our experiments, we test the following hypotheses:
\begin{description}
    \item[H1] Maximizing the product of learned Q-functions (Eq.~\ref{eq:planning-objective}) translates to maximizing long-horizon task success.
    \item[H2] Skills trained with our framework are able to generalize to unseen long-horizon tasks by optimizing Eq.~\ref{eq:planning-objective}.
    \item[H3] Our planning method can be combined with a task planner and UQ (Eq.~\ref{eq:scod-posterior}) to solve TAMP problems.
\end{description}

We evaluate our method on a 3D manipulation domain with 4 skills: $\action{Pick a b}$: pick $\obj{a}$ from $\obj{b}$; $\action{Place a b}$: place $\obj{a}$ onto $\obj{b}$; $\action{Pull a hook}$: pull $\obj{a}$ into the robot's workspace with a $\obj{hook}$; and $\action{Push a hook}$: push $\obj{a}$ with a $\obj{hook}$.

The long-horizon state space $\mathcal{\overline{S}}$ is a sequence of low-dimensional object states that contains information such as 6D poses. 
The policy state spaces $\mathcal{S}^k$ are constructed so that the first $m$ object states correspond to the $m$ arguments of the corresponding skill. 
For example, a state for the policy of $\action{Pick box rack}$ will contain first the $\obj{box}$'s state, then the $\obj{rack}$'s state, followed by a random permutation of the remaining object states.
The policy action spaces $\mathcal{A}^k$ are all 4D. 
For example, a policy-predicted action for $\action{Pick a b}$ specifies the 3D grasp position of the end-effector relative to the target $\obj{a}$ and orientation about the world $z$-axis.

Our evaluation is on 9 different long-horizon tasks (i.e. plan skeletons $\tau$). The tasks cover a range of symbolic and geometric complexities (Fig.~\ref{fig:taps-methods} - Right), with plan skeleton lengths ranging from 4 to 10 skills. 
Each task involves geometric dependencies between actions, which motivates the need for planning. 
We use 100 randomly generated instances (i.e. object configurations) for evaluation on each task.

\subsection{Product of Q-functions approximates task success (\textbf{H1})}

We test \textbf{H1} by comparing STAP to an \textbf{Oracle} baseline that runs forward simulations with policy shooting to find action plans that achieve ground-truth task success. Our method uses learned Q-functions and dynamics to predict task success as the product of Q-functions. We expect that planning with this objective will come close to matching the task success upper bound provided by \textbf{Oracle}.

We compare several planning methods: \textbf{Policy Shooting} and \textbf{Policy CEM}, which use the learned policies to initialize the action sampling distributions (Sec.~\ref{sec:planning-actions}), as well as \textbf{Random Shooting} and \textbf{Random CEM}, which use uniform action priors. 
We also compare with \textbf{Greedy}, which does not plan but greedily executes the skills. 
The evaluation metrics are ground-truth task success, sub-goal completion rate (what percentage of skills in a plan are successfully executed), and predicted task success computed from Eq.~\ref{eq:planning-objective}.

Due to the significant amount of time required to run forward simulations for \textbf{Oracle}, we limit the number of sampled trajectories evaluated during planning to 1000 for all methods. 
This is not enough to succeed at the most complex tasks, and thus, we evaluate on the simplest task from the \texttt{Hook Reach} and \texttt{Constrained Packing} domains.

\begin{figure}
    \centering
    \includegraphics[width=\columnwidth]{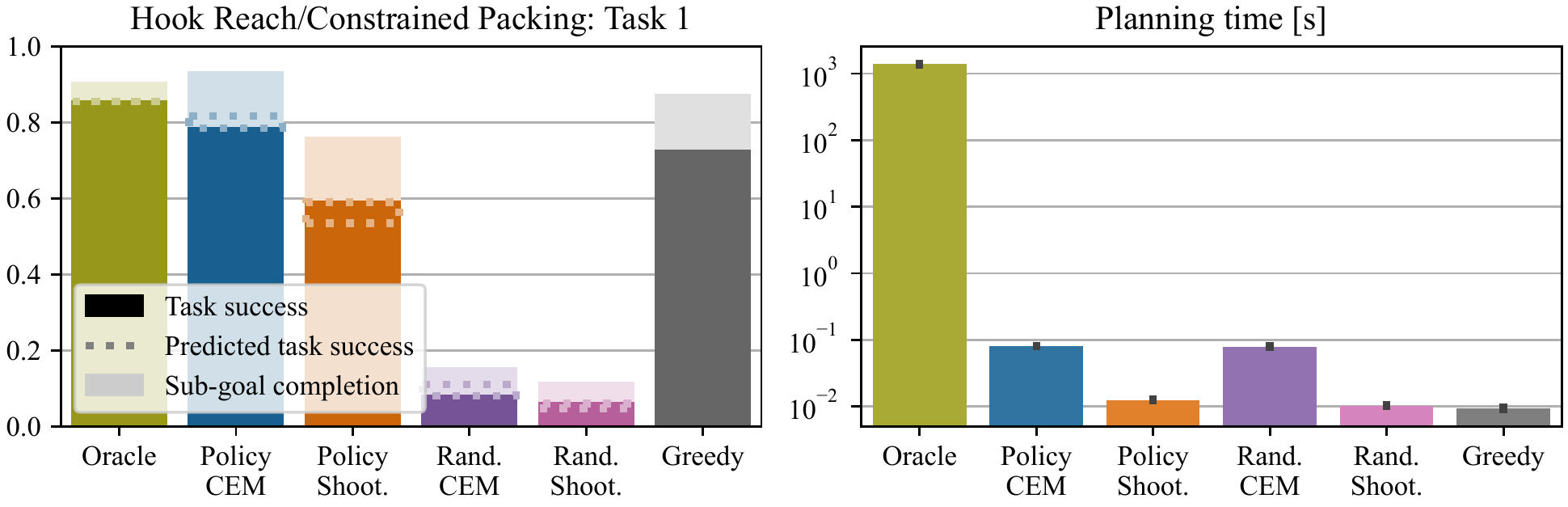}
    \caption{A small-scale experiment comparing the performance of our method to \textbf{Oracle} planning. The left plot shows the average success rates across two domains (\texttt{Hook Reach} and \texttt{Constrained Packing}). The dark bars indicate the ground truth task success, and the light bars indicate sub-goal completion rate, which measures how close the plan was to successfully completing the task. The predicted task success computed from the product of Q-values is indicated by a dotted line. Our method with \textbf{Policy CEM} is able to nearly match the success rate of \textbf{Oracle} while taking 4 orders of magnitude less time, as shown in the plot on the right.}
    \label{fig:exp_a}
\end{figure}

The results from both tasks are averaged and presented in Fig.~\ref{fig:exp_a}. As expected, \textbf{Oracle} achieves the highest success rate, although not perfect because 1000 samples are not enough to solve all of the tasks. \textbf{Policy CEM} nearly matches \textbf{Oracle}'s success rate, which demonstrates that maximizing the product of Q-functions is a good proxy for maximizing task success. \textbf{Policy CEM} also exhibits a low success prediction error, which demonstrates that the learned Q-functions and dynamics generalize well to these unseen long-horizon tasks. Meanwhile, planning with these learned models runs 4 orders of magnitude faster than \textbf{Oracle} and does not require ground-truth knowledge about the environment state or dynamics.

\textbf{Policy Shooting} performs slightly worse than \textbf{Policy CEM}, which demonstrates CEM's strength in finding local maxima through iterative refinement. \textbf{Random CEM} and \textbf{Random Shooting} perform quite poorly, indicating that the planning space is too large (16D for these tasks) for random sampling. \textbf{Greedy} performs strongly, perhaps indicating that these simpler tasks can be solved without planning.

\begin{figure}
    \centering
    \includegraphics[width=\columnwidth]{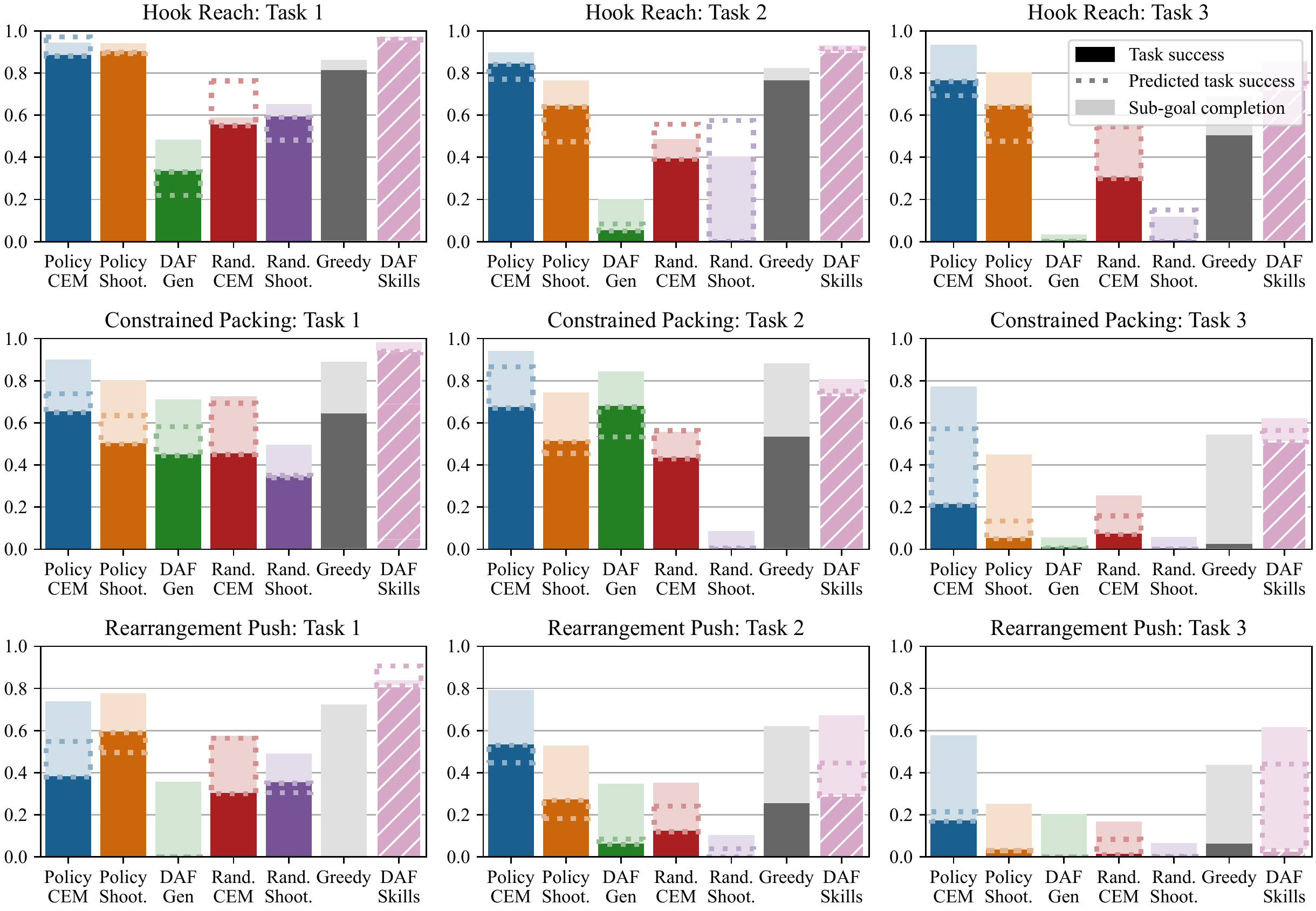}
    \caption{
        Planning experiment with 3 domains, each with 3 tasks. 
        Our method with \textbf{Policy CEM} is able to generalize to all of these tasks without ever seeing them during training.
        On 6 out of the 9 tasks, our method either matches or outperforms \textbf{DAF-Skills}, which is trained directly on the evaluation task. 
        \textbf{DAF-Gen} shows the generalization performance of the \textbf{DAF-Skills} models when evaluated on unseen tasks within the same domain.
    }
    \label{fig:exp_abc}
\end{figure}

\subsection{STAP skills generalize to long-horizon tasks (\textbf{H2})}
\label{sec:experiments-planning}

In this experiment, we test the ability of our framework to solve 9 long-horizon tasks with geometric dependencies between actions. We compare against DAF~\cite{xu2021-daf}, a state-of-the-art method for learning to solve TAMP problems. 
As task planning is outside the scope of this paper, we omit DAF's skill proposal network and compare only to the skills trained with DAF (\textbf{DAF-Skills}), which are comprised of dynamics and affordance models.
DAF's planning objective is similar to ours, except that it evaluates the product of affordances rather than Q-functions. 
We give \textbf{DAF-Skills} the same plan skeleton $\tau$ that is given to our method and augment DAF's shooting planner with CEM for a more even comparison.

Like other model-based RL methods, DAF requires training on a set of long-horizon tasks that is representative of the evaluation task distribution. 
We therefore train one \textbf{DAF-Skills} model per task (9 total) and run evaluation on the same task. 
We also test the ability of these models to generalize to the other two tasks within the same domain (\textbf{DAF-Gen}). 
Since \textbf{DAF-Skills} is trained on its evaluation task, we expect it to perform at least as well as STAP, if not better.
However, we expect \textbf{DAF-Gen} to perform slightly worse than STAP, since the evaluation tasks differ from the training tasks, even if they are similar.
We train all models for 48 hours each and allow $1000$ samples per dimension for planning.

The results are presented in Fig.~\ref{fig:exp_abc}. 
Our method with \textbf{Policy CEM} achieves competitive success rates with \textbf{DAF-Skills} on 4 out of the 9 tasks and outperforms it on 2 tasks with highly complex action dependencies (\texttt{Rearrangement Push}). 
While \textbf{DAF-Gen} matches the performance of \textbf{Policy CEM} on 2 tasks, it gets relatively low success on the others. 
This indicates that skills trained on one long-horizon task may not  effectively transfer to other tasks with similar action dependencies. 
Our method of training skills in independent environments and then generalizing to long-horizon tasks via planning is efficient from a training perspective, since the same trained skills can be used for all downstream tasks.

\subsection{STAP can be extended for TAMP with UQ (\textbf{H3})}
\label{sec:experiments-tamp}
In this experiment, we combine our framework with a PDDL task planner as described in Sec.~\ref{sec:planning-tamp} and evaluate it on two TAMP problems. 
In \texttt{Hook Reach}, the robot needs to decide the best way to pick up a block, which may or may not be in its workspace. 
In \texttt{Constrained Packing}, the robot needs to place a fixed number of objects on the rack but is free to choose among any of the objects on the table. 
To mimic what the robot might find in an unstructured, real-world environment, some of these objects are distractor objects that are initialized in ways not seen by the skills during training (e.g. the blocks can be stacked, placed behind the robot base, or tipped over). 
The task planner may end up selecting these distractor objects for placing on the rack, but since the skills have not been trained to handle these objects, their predicted success (Q-values) may be unreliable. 
UQ is particularly important for such scenarios, so we introduce \textbf{SCOD Policy CEM}, which filters out candidate action plans with high uncertainty in the predicted task success score (Eq.~\ref{eq:planning-objective}).
That is, the $n$ action plans with the highest skill uncertainties (Eq.~\ref{eq:scod-posterior}) are not considered for execution.

\begin{figure}
    \centering
    \includegraphics[width=\columnwidth]{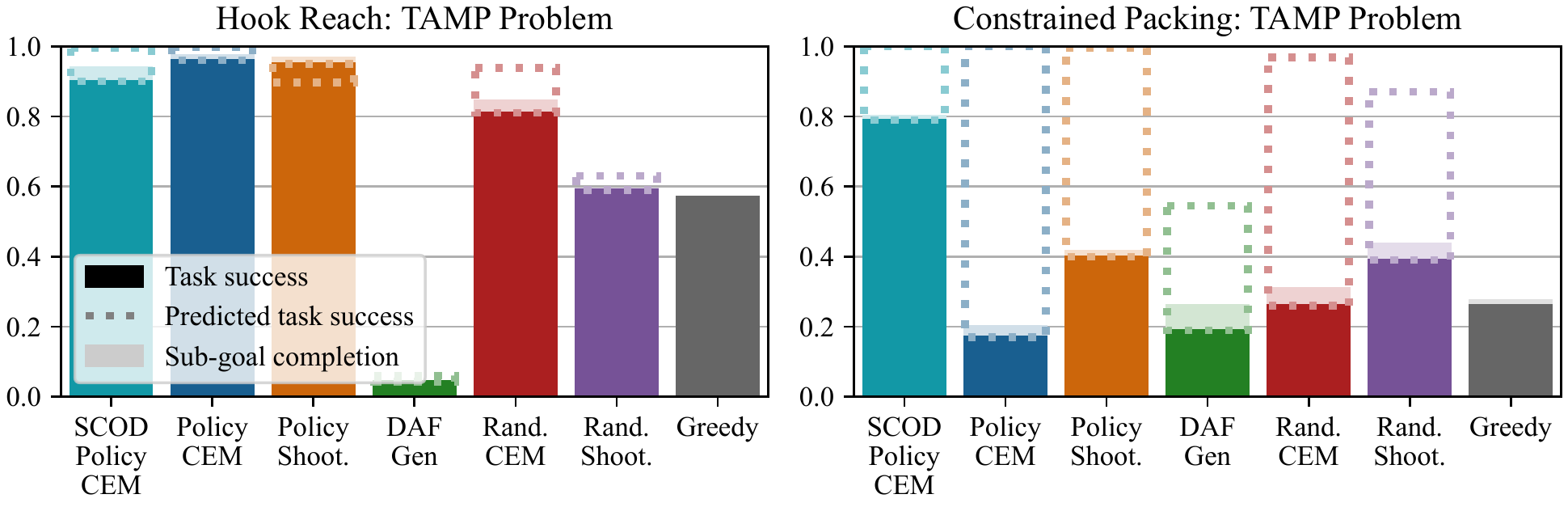}
    \vspace*{-5mm}
    \caption{
        Integration of our planning framework with task planning and UQ to solve TAMP problems. 
        The poor performance of the methods without SCOD in the \texttt{Constrained Packing} problem highlights the importance of UQ when attempting to solve unseen TAMP problems with learned skills.
    }
    \label{fig:tamp_results}
\end{figure}

The results are presented in Fig.~\ref{fig:tamp_results}. \textbf{Policy CEM} achieves 97\% success on the \texttt{Hook Reach} TAMP problem, while \textbf{SCOD Policy CEM} suffers a slight performance drop. However, for \texttt{Constrained Packing}, which contains OOD states, \textbf{SCOD Policy CEM} strongly outperforms the other methods. Exploring different ways to integrate UQ into our planning framework is a promising direction for future work.

\subsection{Real world sequential manipulation}

We demonstrate that skills trained with our framework can be used to perform sequential manipulation tasks in a real robot environment. We take RGB-D images from a Kinect v2 camera and use manually tuned color thresholds to segment objects in the scene. With these segmentations, we estimate object poses using the depth image, which is then used to construct the initial environment state $\overline{s}_1$. Qualitative results are provided in the supplementary video.

\section{Conclusion}
\label{sec:conclusion}
We present a framework for sequencing task-agnostic policies that have been trained independently. The key to generalization is planning actions that maximize the probability of long-horizon task success, which we model using the product of learned Q-values. This requires learning a dynamics model to predict future states and using UQ to filter out OOD states that the skills do not support. The result is a library of skills that can be composed to solve arbitrary long-horizon tasks with complex geometric dependencies between actions. 
Future work includes the investigation of methods for scaling skills to high-dimensional observations, combining the library of learned skills with a set of handcrafted skills, and exploring planning objectives that capture other desirable properties of trajectories, beyond their geometric feasibility.

{\footnotesize
\bibliographystyle{IEEEtranN}
\bibliography{references}
}

\clearpage
\onecolumn

\setlength{\parskip}{1em}

\appendices

\startcontents[sections]
\section*{Appendix -- \textbf{STAP}: Sequencing Task-Agnostic Policies}
The appendix discusses commonly asked questions about our planning framework and provides details on the manipulation skill library used for planning.
Qualitative results and code are made available at \link{https://sites.google.com/stanford.edu/stap/home}{sites.google.com/stanford.edu/stap}.
    
\printcontents[sections]{}{1}

\clearpage
\section{Frequently asked questions}
\label{appx:faq}

\noindent\textbf{Q1}: \textit{What enables our planning framework to generalize to unseen tasks?}

A core hypothesis of this paper is that it is easier to learn \textit{task-agnostic skills} that support long-horizon reasoning than it is to learn a single \textit{long-horizon policy} that can generalize to arbitrary tasks (i.e. skill sequences).
This hypothesis is motivated by two facts: 1) the state-action space associated with all possible long-horizon skill sequences grows exponentially $O(c^H)$ with the length $H$ of the skill sequences considered, where $c$ is some constant; 2) adequately exploring this exponentially sized state-action space during the training of a single long-horizon policy, as required to solve arbitrary tasks, is challenging from both a methodology and engineering standpoint.
In contrast, the state-action that must be sufficiently explored to plan with STAP grows only linearly $O(K)$ with the number of task-agnostic skills $K$ in the skill library.
However, it is essential that STAP's task-agnostic skills are trained (at least in-part) on states that are likely to occur when solving tasks of interest.

Having learned a library of task-agnostic skills, our method plans with the skills at test-time to maximize the feasibility of a specified sequence of skills.
Because the skills have been trained independent of each other and of the planner, every specified sequence of skills can be regarded as an \textit{unseen task} that STAP must generalize to.
Our method accomplishes this via optimization of a planning objective (see Sec.~\ref{sec:taps-grounding}) in a process that involves the skills' policies, Q-functions, and dynamics models. 
Thus, the generality of our method to unseen tasks stems from the compositionality of independently learned skills.

\noindent\textbf{Q2}: \textit{Why is STAP useful for Integrated Task and Motion Planning?}

Task and Motion Planning (TAMP) seeks to solve long-horizon tasks by integrating symbolic and geometric reasoning.
Symbolic task planners and PDDL are commonly used to produce candidate plan skeletons or skill sequences that satisfy a user-specified symbolic goal.
Robotics subroutines, e.g. motion planners, collision checkers, inverse kinematics solvers, are then procedurally invoked to verify the feasibility of the plan skeleton and return a corresponding motion plan.

Different than prior work, STAP presents an avenue to develop general TAMP algorithms centered around learned skills.
While in Sec.~\ref{sec:planning-tamp}, we use STAP to perform geometric reasoning on candidate skill sequences proposed by a symbolic planner, many other instantiations are possible.
For example, success probabilities (Eq.~\ref{eq:planning-objective}) predicted by STAP can serve as a geometric feasibility heuristic to guide task planning with classical search algorithms~\cite{hoffmann2001-ff} or foundation models~\cite{lin2023text2motion}.
Such TAMP algorithms would inherit the efficiency of planning with STAP, as shown in Fig.~\ref{fig:exp_a}.
They would also benefit from the modularity associated with skill libraries, where new skills can be added to support a larger set of tasks and old skills can be updated to improve overall planning performance without the need to modify any other components of the TAMP framework.

\noindent\textbf{Q3}: \textit{What would it take to scale STAP to high-dimensional observation spaces?}

Our framework is not limited to the low-dimensional state space described in Sec.~\ref{sec:experiments}, however, several challenges must be addressed in order to use STAP in conjunction with high-dimensional sensory data such as images or 3D point clouds.
\vspace{-8pt}
\begin{itemize}
    \item \textbf{Skills:} Q-functions must accurately characterize the skill's success probability given the high-dimensional observation. 
    Challenge: the fidelity of skill Q-functions obtained via model-free Reinforcement Learning (RL)~\cite{fujimoto2018-td3,haarnoja2018-sac} may be too low for long-horizon planning. 
    Potential solution(s): acquire skills from large-scale datasets and couple controllable, data-driven learning methods (e.g. offline RL, imitation learning, supervised learning) with data augmentation techniques.
    
    \item \textbf{Dynamics:} The planning state space (Eq.~\ref{eq:long-horizon-domain}) must be amenable to accurate forward prediction over long-horizon skill sequences.
    Challenge: predicted high-dimensional states become coarse over long-horizons~\cite{finn2017-dvf,wu2021greedy} which complicates their use in manipulation planning settings that demand fine-grained geometric detail.
    Potential solution(s): leverage pretrained representations for robotics~\cite{nair2022r3m, karamcheti2023language} and learn latent dynamic models~\cite{hafner2019-dreamer,xu2021-daf} for forward prediction.
\end{itemize}
\vspace{-8pt}
Since STAP is reliant on the quality of the underlying skill library, we expect the capabilities of our method to improve with advancements in robot skill acquisition and visuomotor policies, representation learning, and video prediction for robotics.

\noindent\textbf{Q4}: \textit{How else can uncertainty quantification be incorporated into planning?}

In our TAMP experiments (Sec.~\ref{sec:planning}), we take a filtering-based approach to robustify STAP planning in out-of-distribution scenarios.
Specifically, we use Sketching Curvature for Out-of-Distribution Detection (SCOD)~\cite{sharma2021-scod} for uncertainty quantification (UQ) of Q-functions and disregard the $n$ plans with the most uncertain Q-values at each planning iteration.

While SCOD imposes no train-time dependencies on any algorithms in our framework, its forward pass is computationally and memory intensive and slows planning as a result.
For faster planning, UQ alternatives such as deep ensembles~\cite{ensemblereview2021} or Monte-Carlo dropout~\cite{dropoutunc2016} can be employed which, in contrast to SCOD, require modifying the algorithms used to learn skills. 
We further note that the described filtering-based optimization approach can be substituted with more sophisticated planning techniques, several of which have been implemented and verified to work with STAP. 
For example, we could formulate a distributionally robust variant of our planning objective (Eq.~\ref{eq:planning-objective}) to optimize a lower-confidence bound of the Q-values:
\begin{equation*}
    J(a_{1:H}; \overline{s}_1)%
        = \E{\overline{s}_{2:H} \sim \overline{T}_{1:H-1}}{\Pi_{h=1}^{H} \func{Q_h}{\!\func{\Gamma_h}{\overline{s}_h}, a_h} - \alpha F_\text{unc}(\overline{s}_h, a_h; Q_h, \Gamma_h)},
\end{equation*}
where $F_\text{unc}$ quantifies the uncertainty of Q-function $Q_h$ evaluated at $s_h=\Gamma(\overline{s}_h)$ and $a_h$, and $\alpha$ is a scalar hyperparameter (e.g. the z-score). 
The uncertainty function $F_\text{unc}$ is determined by the choice of UQ method.
For deep ensembles, this would correspond to the empirical variance of the ensemble predictions $F_\text{unc}(\overline{s}_h, a_h; Q_h, \Gamma_h) = \V{}{Q_h(\Gamma_h(\overline{s}_h), a_h)}$.
Other choices of $F_\text{unc}$ include posterior predictive variances provided by SCOD or coherent risk measures such as Conditional Value at Risk (CVaR)~\cite{brown2020bayesian}. 
We have experimetally validated such formulations of STAP, and ultimately, the appropriate choice of planning objective and optimization scheme should correspond to the evaluation tasks and domains of interest.

\noindent\textbf{Q5}: \textit{What are the main limitations of our method?}

Currently, our method is limited in settings with high degrees of partial observability and stochastic dynamics.
Examples include mobile manipulation in unknown environments and interacting with dynamic agents, where it may be intractable to predict future states necessary for look-ahead planning.
While these tasks are beyond the scope of this work, we note that our method is agnostic to the specific choice of skills and dynamics models, and thus, components that account for stochasticity could be used interchangeably. 
In this work, we focus on generalization to geometrically challenging manipulation problems.

\clearpage
\section{Manipulation skill library}
\label{appx:skill-library}

\subsection{Parameterized manipulation primitives} 
All variants of STAP interface with a library of manipulation skills $\mathcal{L}=\{\psi^1, \ldots, \psi^K\}$.
Each skill $\psi$ consists of a learned policy $\func{\pi}{a \given s}$ and a parameterized manipulation primitive~\cite{felip2013manipulation} $\phi(a)$.
The policy is trained to output parameters $a \sim \func{\pi}{a \given s}$ that results in successful actuation of the primitive $\phi(a)$ in a contextual bandit setting (Eq.~\ref{eq:skill-mdp}) with a binary reward function $R(s, a, s')$.
Our library consists of four skills, $\mathcal{L}=\{\psi^{\text{Pick}}$, $\psi^{\text{Place}}$, $\psi^{\text{Pull}}$, $\psi^{\text{Push}}\}$, used to solve tasks in simulation and in the real-world.
The policies must learn to manipulate objects with different geometries (e.g. $\pi^{\text{Pick}}$ is used for both $\actioncall{Pick}{box}$ and $\actioncall{Pick}{hook}$).  
We describe the parameterization and reward function of each skill below.
A reward of $r = 0$ is provided if any collision occurs with a non-argument object.
For example, if $\psi^{\text{Place}}$ collides with \textit{rack} while executing $\actioncall{Place}{box}{table}$.
\vspace{-8pt}
\begin{itemize}
    \item \textbf{$\actioncall{Pick}{obj}$}: the parameter $a$ represents the grasp pose of \textit{obj} in the coordinate frame of \textit{obj}. The policy $\pi^{\text{Pick}}$ receives a reward of $r = 1$ if the primitive $\phi^{\text{Pick}}$ successfully grasps and picks up \textit{obj}.
    \item \textbf{$\actioncall{Place}{obj}{rec}$}: the parameter $a$ represents the placement pose of \textit{obj} in the coordinate frame of \textit{rec}. The policy $\pi^{\text{Place}}$ receives a reward of $r = 1$ if the primitive $\phi^{\text{Place}}$ places \textit{obj} stable atop \textit{rec}.
    \item \textbf{$\actioncall{Pull}{obj}{tool}$}: the parameter $a$ represents the initial position, direction, and distance of a pull on \textit{obj} with \textit{tool} in the coordinate frame of \textit{obj}. The policy $\pi^{\text{Pull}}$ receives a reward of $r = 1$ if the primitive $\phi^{\text{Pull}}$ moves \textit{obj} toward the robot by a minimum of $0.05m$. 
    \item \textbf{$\actioncall{Push}{obj}{tool}{rec}$}: the parameter $a$ represents the initial position, direction, and distance of a push on \textit{obj} with \textit{tool} in the coordinate frame of \textit{obj}. The policy $\pi^{\text{Push}}$ receives a reward of $r = 1$ if the primitive $\phi^{\text{Push}}$ moves \textit{obj} away from the robot by a minimum of $0.05m$ and if the final pose of \textit{obj} is underneath \textit{rec}.
\end{itemize}

\subsection{Training manipulation skills}
We use the Soft Actor-Critic~\cite{haarnoja2018-sac} (SAC) algorithm with original hyperparameters to simultaneously learn a stochastic policy $\func{\pi}{a \given s}$ and Q-function $Q^\pi(s, a)$ for each skill $\psi$. 
All models are trained for $200k$ single-step episodes and the $(s, a, s')$ transitions are stored in a replay buffer $\mathcal{D}^\psi=\{(s^i, a^i, s'^i)\}_{i=1}^{200k}$ for each skill $\psi$.
The replay buffer data is later used to train the dynamics models $\func{\overline{T}^k}{\overline{s}, a^k}$ (Sec.~\ref{sec:training-dynamics}) and calibrate the weights $w^k$ used by SCOD for UQ (Sec.~\ref{sec:training-scod}).

\end{document}